# Attention Residual U-Net for Enhanced Cross-Dataset Polyp Localization in Endoscopy

Romil Imtiaz
*Department of Computer Science and Biomedical Informatics*
*University of Thessaly*
Lamia, Greece
rimtiaz@uth.gr

Dimitris K. Iakovidis
*Department of Computer Science and Biomedical Informatics*
*University of Thessaly*
Lamia,Greece
diakovidis@uth.gr

***Abstract*— Precise localization of pathologies in endoscopy is a critical task in medical imaging, with deep learning (DL) frameworks now serving as the primary methodology for automated lesion detection. Today, generic object detection methods, such as YOLO, provide state-of-the-art performance in this context. However, their robustness often decreases under real-world clinical conditions due to variations across endoscopic devices, intestinal content, dataset bias, and the presence of small or low-contrast lesions. Motivated by these challenges, we propose a novel polyp detection framework based on a Attention Residual U-Net offering enhanced polyp localization. Instead of directly regressing bounding boxes from global feature representations, the proposed method learns lesion-focused spatial representations and converts them into bounding-box predictions during inference. Unlike conventional encoder–decoder localization frameworks that directly propagate all encoder activations into the decoding pathway, the proposed method selectively suppresses irrelevant background responses before feature reconstruction. This encourages the network to focus on lesion-relevant spatial evidence under heterogeneous imaging conditions. To achieve this, the framework integrates residual feature propagation, attention-guided encoder filtering, and multi-scale contextual learning. The proposed framework was evaluated on multiple public endoscopic datasets acquired using different devices. The comparative experimental results validate its effectiveness over the relevant state-of-the-art methods, with an improvement up to 22.0% in terms of the area under the receiving operating characteristic (AUC).***

***Keywords—Polyp detection, attention, residual learning, cross-dataset evaluation, deep Learning***

## I. INTRODUCTION

Polyp localization in endoscopic imaging is an important problem in medical computer vision because colorectal polyps are considered precancerous lesions that may develop into colorectal cancer if not detected and removed at early stages. Accurate localization of abnormal tissue regions during endoscopic examination therefore plays a critical role in early diagnosis and clinical decision support. However, false negative detections may delay diagnosis and increase clinical risk, motivating the development of robust automated localization systems for challenging endoscopic imaging conditions.

Single stage detection models, such as You Only Look Once (YOLO), can be both efficient and accurate, e.g., YOLOv7 [1], YOLOv8, and YOLOv11 [2] models have been successfully applied to various object detection tasks in medical imaging, Autonomous driving, robotics, agriculture, surveillance, and manufacturing. However, these models are prone to errors in cases of noisiness, illumination variations, and complicated background in images. Endoscopic images usually have polyps of small sizes, unusual shapes, and color similar to background tissues. Thus, these objects are hard to identify with standard feature representations.

Recent polyp localization approaches have explored attention-guided feature refinement and multiscale contextual representation to improve lesion localization under challenging imaging conditions. Attention mechanisms can help suppress irrelevant background responses and emphasize lesion-related spatial regions [3], while multiscale representation learning improves the detection of polyps with `varying sizes and [4]. However, despite the success of these models, the problem of distinguishing lesion from background is hard since lesions might appear anywhere within an image.

An important limitation in many existing object localization frameworks is the uncontrolled propagation of irrelevant feature activations during hierarchical feature extraction. In endoscopic imaging, background structures, such as mucosal folds, lumen boundaries, specular reflections, and illumination artifacts may generate strong activations that compete with lesion-related representations. When these irrelevant responses are directly propagated through encoder-decoder feature pathways, they may interfere with lesion localization and reduce robustness under heterogeneous imaging conditions. This limitation becomes particularly critical during cross-dataset evaluation, where variations in acquisition conditions and image appearance may further amplify unstable background activations.

Motivated by this challenge, this paper proposes a novel framework that explicitly suppresses irrelevant encoder responses before feature reconstruction to encourage lesion-focused spatial representation learning under challenging endoscopic conditions. It is based on DL architecture incorporating a detector that deals with the problem of irrelevant feature propagation and pays more attention to polyps. To reduce the impact of irrelevant background features and pay more attention to lesions, we developed a special attention guided module. Such approach allows us to filter out background information and obtain more relevant features for lesion detection. The proposed framework also incorporates residual feature learning to improve optimization stability and feature propagation, as well as multiscale contextual representation learning to better capture lesions with varying sizes and appearance. In addition to attention mechanism, the developed architecture uses residual learning and multiscale features extraction, which are particularly helpful in endoscopy detection tasks, since polyps are hard to detect due to various variations in their sizes, appearance, and conditions of acquisition. Furthermore, the proposed

This project has received funding from the European Union's HORIZON TMA MSCA Doctoral Networks (HORIZON-MSCA-2023-DN-01) under Grant Agreement n°101169012 (INTELLI-INGEST Project).

Author version of the accepted paper for publication in the 2026 IEEE Conference on Computational Intelligence in Bioinformatics and Computational Biology (CIBCB).

framework learns lesion-focused spatial representations that are transformed into bounding-box localization outputs during inference, enabling compatibility with detection-oriented evaluation protocols. A cross-dataset evaluation setting was used to assess and validate the advantages of the proposed framework.

The rest of this paper consists of four sections. Section II provides a review of relevant state-of-the-art methods, and section III describes the proposed framework and its methodological details. The results of the experimental evaluation are presented in section IV and the conclusions of this study are summarized in the last section.

## II. Related Works

Deep learning-based polyp detection has become a key area in computer-aided endoscopy, as it can support clinicians by automatically identifying suspicious lesion areas in colonoscopy frames. Existing detection approaches include two-stage detectors, single-stage detectors, transformer-based models, and region-aware encoder–decoder frameworks [5], [6]. Earlier object detection methods, such as Faster R-CNN introduced region proposal mechanisms for object localization [7], while single-stage detectors such as Single Shot MultiBox Detector (SSD) and YOLO improved detection efficiency by predicting object locations directly from feature maps [2], [8]. In endoscopic applications, YOLO-based models are widely used because of their speed and deployment simplicity, with YOLOv7 introducing a trainable bag-of-freebies strategy for real-time detection [2]. Recent YOLO-based frameworks, including YOLOv8 and YOLOv11, have also been used in medical detection tasks due to their practical deployment benefits [2]. For polyp detection specifically, Lalinia and Sahafi [9] applied YOLOv8 to colonoscopy images and demonstrated efficient localization performance on public polyp datasets.

Recent detection-oriented approaches have attempted to improve robust polyp localization through stronger multiscale feature fusion, contextual representation learning, and enhanced bounding-box regression strategies [10], [11]. These methods improve detection of polyps with varying sizes and appearances by enhancing feature aggregation and localization refinement mechanisms. Other approaches have incorporated contextual receptive field enhancement and refined multiscale pooling operations to improve contextual representation learning for gastrointestinal polyp localization [11]. Additional studies have further explored multiscale feature integration strategies to improve localization robustness under complex polyp appearance variations [12].

Despite these improvements, previous detection methods still show limitations in difficult endoscopic conditions. Existing approaches mainly improve detection heads, feature fusion, or multiscale representation, but they do not explicitly address the propagation of irrelevant background activations before localization. As a result, small or low-contrast polyps may be missed, and visually similar structures such as mucosal folds, specular reflections, bubbles, and lumen boundaries may contribute to false-positive detections. This limitation becomes more evident under cross-dataset evaluation, where changes in illumination, acquisition device, and tissue appearance can amplify unstable background responses. Motivated by these limitations, the proposed framework introduces attention-guided region-aware feature learning to suppress irrelevant encoder activations before localization.

Attention mechanisms have also been studied to reduce the influence of irrelevant background information. In endoscopic images, not all extracted features contribute meaningfully to detection. Lumen boundaries, tissue folds, and specular reflections can create strong activations that interfere with accurate lesion localization. Xie *et al.* proposed an attentional feature fusion strategy for gastric polyp detection, using attention to suppress background noise and emphasize useful target features [13]. Recent attention-based modifications to detection frameworks have shown that feature selectivity can be improved while maintaining computational efficiency in colorectal polyp detection [14]. These findings support the use of attention-guided feature refinement in proposed work, where the objective is to filter out irrelevant encoder features while preserving lesion-related information for detection.

Encoder–decoder frameworks have also been explored for polyp analysis because they can preserve spatial details while learning lesion-aware representations. Several studies have introduced attention-based refinement, multiscale contextual learning, and progressive region enhancement to improve polyp localization under complex colonoscopy conditions [15], [16], [17]. These works demonstrate the value of spatial feature refinement; however, detection-oriented localization remains challenging when irrelevant background activations are propagated before bounding-box generation. Despite successful results in various medical image analysis problems, direct implementation of U-Net architectures for the task of localization leads to certain limitations. First of all, conventional architectures of U-Nets [18] use the mechanism of direct skip connections, in which features from the encoder part are transferred into the decoder. Although such an approach helps to preserve spatial details, it also allows activations caused by non-relevant areas to pass through: specular reflections, lumen borders, folds of tissue, and noise. The problem arises since those areas can be confused with lesion regions during endoscopic imaging.

Secondly, the issue related to the receptive field of conventional convolutions limits the possibilities of using standard U-Nets in endoscopic localization. The problem is that standard convolutions use fixed-size spatial kernels; thus, there is no way to capture both local lesion texture and its surroundings in one network. Such a restriction becomes obvious when considering the diversity of the scale at which polyps are found in endoscopy. It means that small lesions require local texture representations while large lesions need broader contextual understanding to be localized.

Thirdy, optimization of deeper architectures tends to be problematic due to gradient vanishing. Deeper networks suffer from unstable learning and weak convergence and generalization performance. These restrictions encourage the use of residual learning, attention-based feature refinement, and multi-level contexts in the proposed architecture.

These limitations become even more critical under cross-dataset evaluation settings, where variations in illumination conditions, acquisition devices, image quality, and tissue appearance may amplify irrelevant feature activations and reduce localization stability. When models rely heavily on dataset-specific background patterns, their performance may degrade significantly under unseen imaging conditions. The proposed framework attempts to address this limitation by suppressing irrelevant encoder activations before feature reconstruction and encouraging lesion-focused spatial representation learning. By reducing the influence of unstable background responses and integrating multiscale contextual representations, the proposed architecture is expected to improve robustness under heterogeneous endoscopic imaging conditions and cross-dataset domain variation.

Beyond attention, cross-dataset robustness has become an increasingly important issue. Models trained on one dataset often show reduced performance when evaluated on data from different institutions or acquisition environments. Ma et al. introduced the LDPolyp Video benchmark, a large-scale colonoscopy video dataset containing diverse polyp appearances, enabling evaluation under more realistic visual variability [19]. Recent studies on generalisability have also shown that model performance can differ considerably across datasets, highlighting the need for evaluation beyond a single-source dataset [20]. Motivated by these findings, we evaluate the proposed framework under a source-to-target cross-dataset setting, where the model is trained on a source endoscopic dataset and tested on multiple unseen target datasets without fine-tuning, such as retraining the final layers, updating the full network weights, or adapting the model using samples from the target datasets before evaluation. Compared with direct YOLO-style detection, our method incorporates region-aware feature learning, attention-guided filtering, and multiscale contextual modeling to improve localization under noisy and heterogeneous imaging conditions, while still producing bounding-box outputs for detection evaluation.

## III. METHODOLOGY

This work proposes a detection-oriented localization framework for robust polyp detection in challenging endoscopic images. The architecture of the proposed framework is shown in Fig. 1. This network is composed of an encoder-decoder framework supplemented by residual feature learning, attention-based feature refining, and multiple context modeling mechanisms. Instead of regressing bounding boxes from global features in the manner typical for traditional object detection systems, the proposed model learns lesion-specific spatial activations and translates them to bounding box predictions during testing. In this way, we are able to retain fine details while suppressing any noise responses often seen in endoscopic images. Specifically, an encoder learns hierarchical visual features from the input image, while a decoder refines them into lesion-specific spatial activations utilizing attention-based skip connections.

To boost robustness in dealing with varying lesion sizes and appearances, a multi-scale bottleneck layer is inserted between the encoder and decoder modules. As a result, the final output of the system is a lesion confidence map from which bounding box predictions are generated via post-processing. By combining selective feature propagation with multi-scale contextual learning, the proposed framework aims to improve localization reliability under noisy and heterogeneous imaging conditions.

### A. Residual Feature Learning

Aiming to enhance the robustness and feature flow through the optimization process, residual convolutional units are utilized in the design of the proposed architecture's encoder, shown in Fig. 2. Rather than making the CNN learn a new transformation for every single layer, the residual learning concept enables the model to learn the residual mapping of the initial data:

$$y = f(x) + x \tag{3}$$

where $x$ represents the input feature map and $f(x)$ denotes the learned residual transformation. It helps improve gradient propagation and deep feature representation learning. With residual learning, there is a chance for efficient feature reuse by ensuring that low-level information is maintained through high-level feature refinement. In difficult endoscopic imagery, it is hard to learn the difference in texture between the lesion tissue and mucosa.

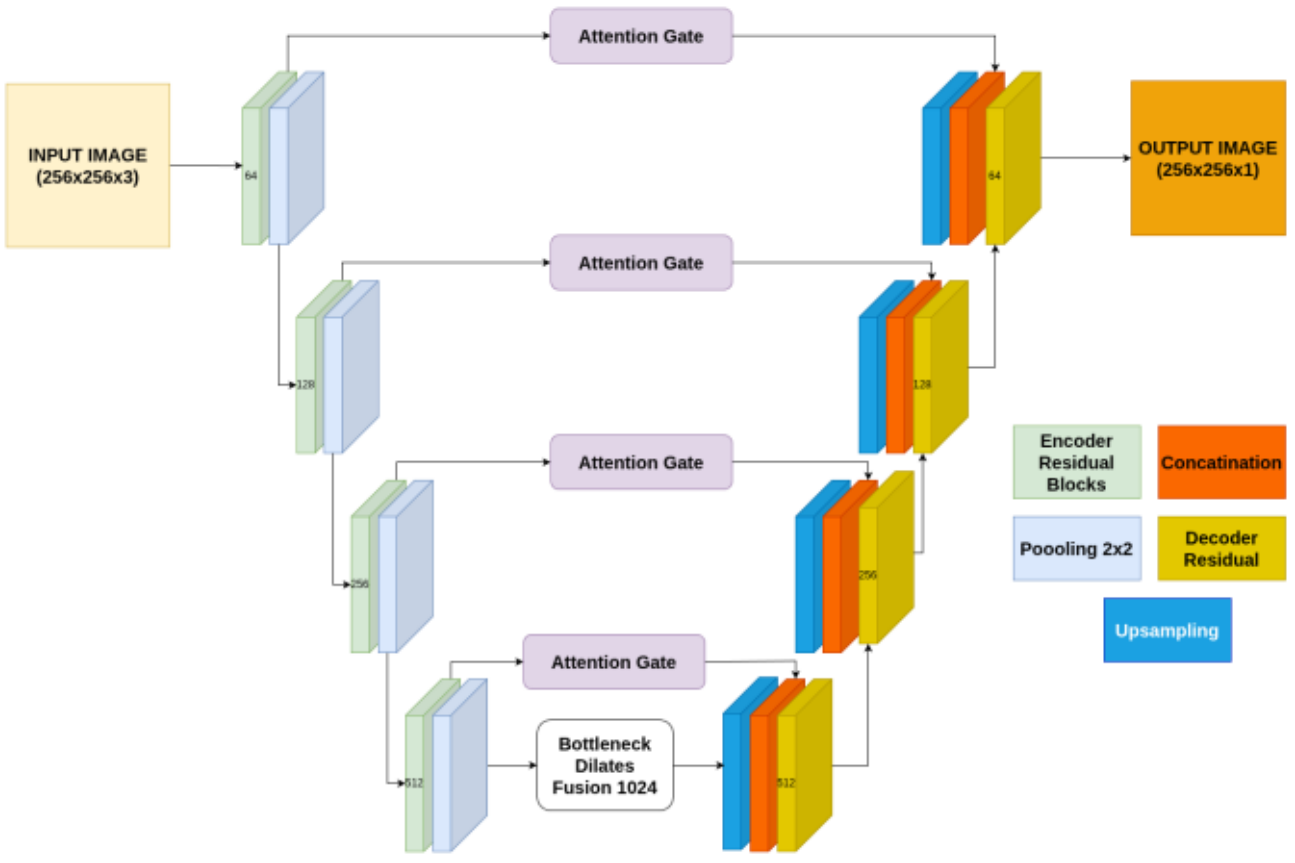

Fig. 1. Overall architecture of the proposed detection-oriented Attention Residual U-Net with residual encoding, attention-guided feature refinement, and multi-scale contextual learning.

Without the aid of residual learning, more advanced feature extraction layers would likely have been affected by unstable optimization and poor convergence properties. This would compromise the capacity of the network to extract discriminatory lesion features, especially when dealing with imaging artifacts. With the inclusion of residual connections, the proposed model addresses this challenge.

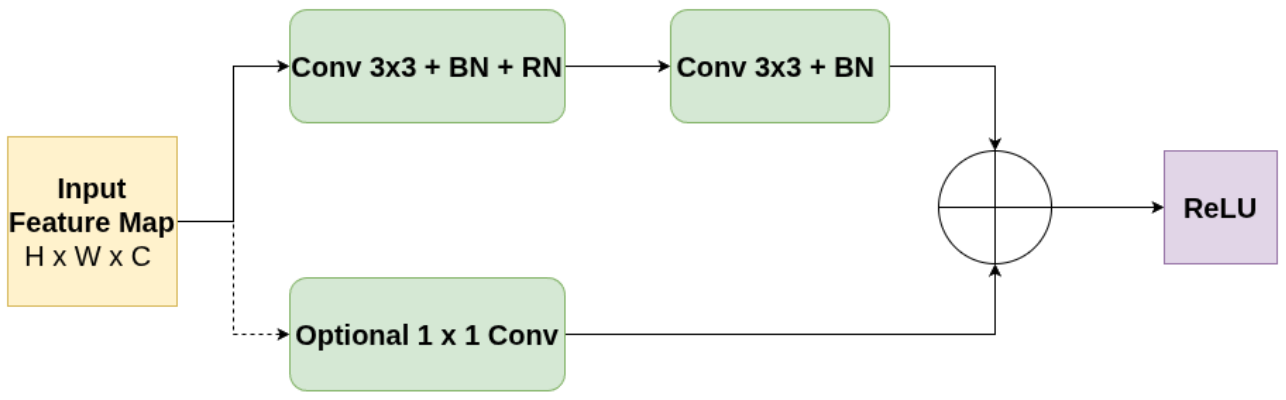

Fig. 2. Attention gate mechanism used to suppress irrelevant encoder features before decoder feature fusion. The circled plus symbol denotes element-wise addition of transformed feature maps.

### B. Attention-Guided Feature Suppression

One of the key contributions of the proposed framework is the introduction of attention-guided feature suppression within the skip connection pathway. The attention mechanism used in the proposed architecture is illustrated in Fig. 3. Instead of directly forwarding all encoder features to the decoder, the proposed model selectively filters encoder activations before feature fusion.

Given encoder features $x$ and decoder guidance features $g$, the attention coefficient is computed as

$$\alpha = \sigma\left(W_\psi\left(\mathrm{ReLU}\left(W_x x + W_g g\right)\right)\right) \tag{4}$$

where $x$ denotes the encoder feature map transferred through the skip connection, $g$ denotes the decoder guidance feature map, $W_x$ and $W_g$ are learnable linear transformations implemented using 1×1 convolutions, $W_\psi$ is a learnable projection used to generate the attention response, $\mathrm{ReLU}(\cdot)$ is the nonlinear activation function, $\sigma(\cdot)$ denotes the sigmoid function, and α represents the spatial attention coefficient. . The refined feature representation is then obtained as:

$$x' = \alpha \cdot x \quad (5)$$

where $x'$ denotes the attention-refined encoder feature map. This mechanism enables the decoder to guide the selection of relevant encoder information by suppressing activations that are unlikely to correspond to lesion regions. In endoscopic imagery, background structures such as folds, illumination artifacts, and specular highlights often produce strong feature responses that may negatively influence localization performance.

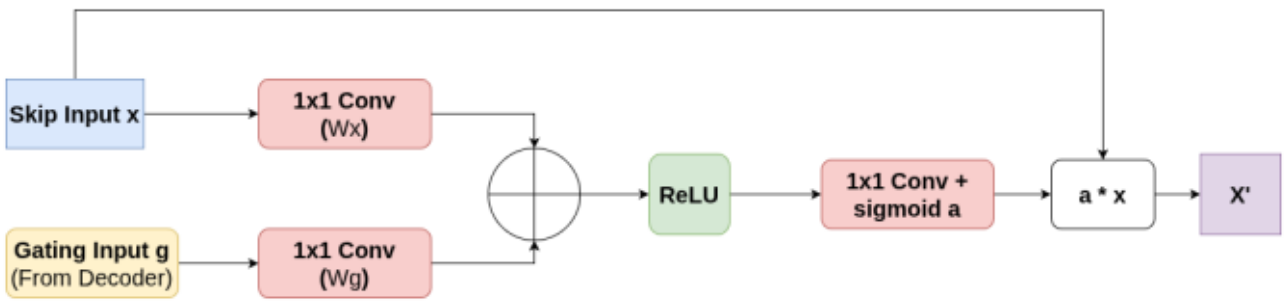

Fig. 3. Attention gate mechanism used to suppress irrelevant encoder features before decoder feature fusion.

Without attention-guided filtering, irrelevant activations are directly propagated into the decoder pathway, which can increase false positive predictions and reduce localization precision. By suppressing background interference before feature fusion, the proposed framework encourages the network to focus on lesion-relevant spatial patterns, improving robustness under visually complex imaging conditions.

## C. Multi-Scale Contextual Representation

To improve robustness against variations in lesion size and appearance, a multi-scale bottleneck module is introduced between the encoder and decoder stages, as shown in Fig. 4. The module applies parallel dilated convolutions using multiple receptive fields:

$$F_r = \mathrm{Conv}_{3\times3}^{(r)}(x) \quad (6)$$

where $r$ represents the dilation rate and $F_r$ denotes the feature map generated by the dilated convolution branch with dilation rate $r$. Thus $F_1, F_2$, and $F_3$ represent feature maps produced using dilation rates 1, 2, and 3, respectively. The resulting feature maps are fused as:

$$F = \phi([F_1, F_2, F_3]) \quad (7)$$

where $[.]$ denotes feature concatenation and $\phi$ represents feature fusion using convolutional projection.

This design allows the network to simultaneously capture fine local texture information and broader contextual representations. Small lesions benefit from local receptive fields, while larger lesions require wider contextual understanding. By integrating multiple receptive fields within the bottleneck representation, the proposed framework improves localization robustness across different lesion scales.

Without multi-scale contextual learning, the network may struggle to detect small lesions or incorrectly interpret surrounding anatomical structures. The proposed multi-scale module therefore improves the ability of the model to adapt to heterogeneous lesion appearance and spatial variability.

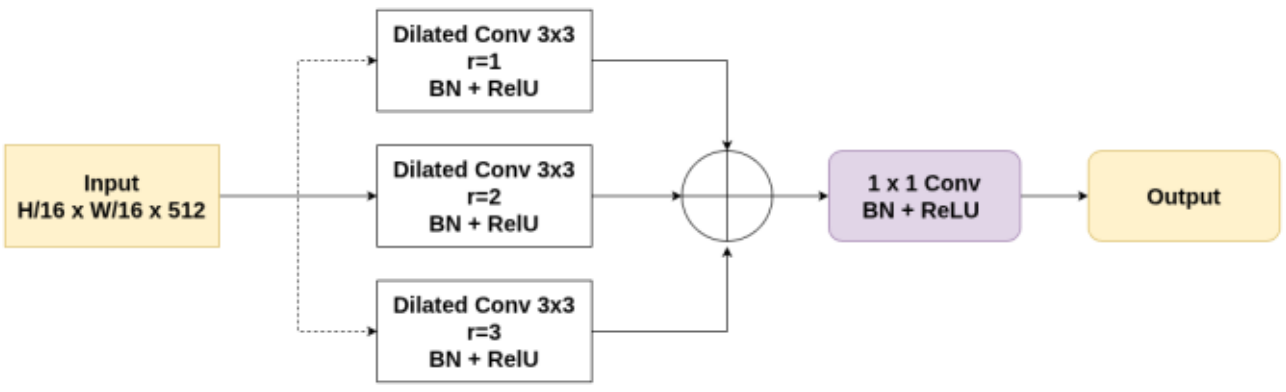

Fig. 4. Multi-scale bottleneck module using parallel dilated convolutions for contextual feature extraction at multiple receptive fields. The circled plus symbol denotes feature-map fusion by addition before the final 1×1 convolution.

## D. Bounding Box Generation

Although the proposed framework learns spatial lesion representations, the final objective of the study remains detection-oriented localization. Therefore, the predicted confidence map is transformed into bounding-box predictions during inference. During inference, the predicted sigmoid probability map is resized to the original image resolution and threshold using a fixed value of 0.5 to generate binary lesion regions. Pixels satisfying $P(i,j) \geq 0.5$ are assigned to the lesion region, while remaining pixels are treated as background. The same threshold value was used for all validation, internal test, and external cross-dataset evaluations without dataset-specific tuning. he final bounding box is generated by identifying all positive pixels in the binary prediction map and computing the minimum axis-aligned rectangle enclosing the detected lesion region:

$$B_k = [x_{min}, y_{min}, x_{max}, y_{max}] \quad (8)$$

where $x_{max}, x_{min}$ denote the minimum and maximum horizontal coordinates of positive pixels, while $y_{max}, y_{min}$ denote the corresponding vertical coordinates. If no positive pixels are detected, no bounding box is generated for that image. This post-processing stage creates a direct connection between region-aware feature learning and standard detection evaluation. By converting spatial predictions into bounding boxes, the proposed framework remains compatible with conventional detection metrics and enables direct comparison with YOLO-based detectors.

Without this conversion process, the framework would produce only spatial confidence responses, making comparison with bounding-box detection methods less consistent. The proposed bounding-box generation strategy therefore preserves the advantages of region-aware learning while maintaining compatibility with detection-oriented evaluation protocols.

# IV. Results and Discussion

Sensitivity, specificity, area under the curve (AUC) [21], and F1 [22] score are metrics utilized for validating the proposed framework. The sensitivity [23] reflects the capability of the model for lesion detection, whereas the specificity [24] reflects the ability of the model to suppress false positives. In contrast, the F1 score shows the trade-off between the precision and recall rates. The AUC reflects the discriminative capability of the model. This metric is crucial for performing endoscopic lesion detection due to the more complicated conditions for lesions detection and high false positives.

To evaluate the polyp localization performance of the proposed framework, the LDPolyp [25] was used as the source dataset for model training and internal testing, while KUMC [26], EDD2020[27], and Kvasir-SEG[8] were used as external datasets for testing the trained system with different imaging conditions, lesion appearances, and data distributions. The details of the datasets are summarized in Table I. LDPolyp was selected as the source dataset because it provides a substantially larger number of annotated polyp frames compared with the external datasets, making it more suitable for training deep lesion localization models. None of the images of the external testing sets was used during training or model selection.

For comparison YOLOv7 and YOLOv11 were included, since they have been used by state-of-the-art polyp detection approaches (section II). This comparison enables to assess whether the proposed attention-guided and multi-scale localization framework provides benefits beyond standard YOLO-based detection pipelines.

Each image is associated with annotations represented in YOLO format, where each object is defined using the normalized center coordinates and dimensions of the bounding box. The annotation representation is defined as

$$B = (c_x, c_y, w, h) \quad (1)$$

where $(c_x, c_y)$ denote the center coordinates of the bounding box, while $w$ *and* $h$ represent the normalized width and height of the lesion region.

All experiments were conducted using a frame-level partitioning strategy on the LDPolyp dataset. The proposed Attention Residual U-Net was trained using the Adam optimizer with an initial learning rate of $1 \times 10^{-4}$. The loss function combined binary cross-entropy and Dice loss. Input images were resized to 256×256 pixels, the batch size was set to 8, and training was performed for up to 80 epochs. Model selection was performed using the best validation performance, while early stopping monitored validation loss with a patience of 15 epochs. The learning rate was reduced by a factor of 0.5 when the validation loss did not improve for 5 epochs.

TABLE I. DATASETS USED IN THE EXPERIMENTS.

| Dataset | Usage | Purpose | Polyps Images | Normal |
|---|---|---|---|---|
| LDPolyp | Model Training/ Reference Teststing | Model training | 33,884 | 6,382 |
| KUMC | External test set | Cross-dataset evaluation | 103 | 30 |
| Kvasir-SEG | External test set | Cross-dataset evaluation | 1000 | 35 |
| EDD 2020 | External test set | Cross-dataset evaluation | 127 | 53 |

### *A. ReferenceResults on LDPolyp Dataset*

The focus of the experimental evaluation was on cross-dataset assessments; however, for reference purposes we performed an additional 5-fold cross-validation experiment on non-overlapping subsets of the LDPolyp dataset. The dataset was partitioned at the frame level, with 20% of the samples used for validation and 10% for testing in each fold. The reported results correspond to the average performance obtained across all five folds. The proposed framework achieved the highest overall polyp detection performance among the compared methods, obtaining an AUC of 97.0% and an F1-score of 98.6%. In comparison, YOLOv7 achieved an AUC of 79.5% and YOLOv11 achieved an AUC of 78.4%. The proposed framework also produced higher specificity compared with both YOLO-based detectors, indicating improved suppression of irrelevant background activations.

The enhanced performance on the LDPolyp dataset suggests that the proposed architecture successfully learned discriminative lesion representations that generalize beyond the training samples. The attention-guided feature refinement mechanism appears to reduce the propagation of irrelevant encoder activations originating from specular reflections, tissue folds, and illumination artifacts. This selective feature filtering likely contributed to the reduction of false positive detections observed in the proposed framework.

In addition, the integration of residual feature learning improved optimization stability during training. The residual encoder blocks enabled more reliable gradient propagation across deeper layers, which improved feature representation learning under visually complex endoscopic conditions. The multi-scale contextual bottleneck further improved localization performance by allowing the network to simultaneously capture fine lesion textures and broader anatomical context.

TABLE II. RESULTS ON LDPOLYP DATASET.

| Methods | Sensitivity | Specificity | AUC | F1-score |
|---|---|---|---|---|
| YOLOv7 | 91.9 | 67.0 | 79.5 | 91.5 |
| YOLOv11 | 94.4 | 62.4 | 78.4 | 91.5 |
| **Proposed** | **98.5** | **95.4** | **97.0** | **98.6** |

### *B. Cross-Dataset Evaluation on KUMC*

Table III presents the cross-dataset evaluation results obtained on the KUMC dataset. All models experienced performance degradation compared with the LDPolyp evaluation results, which reflects the presence of domain variation between datasets. Differences in acquisition devices, illumination conditions, tissue appearance, and imaging protocols can substantially influence feature distribution and therefore reduce generalization performance.

TABLE III. POLYPS DETECTION (%) ON KUMC DATASET.

| Methods | Sensitivity | Specificity | AUC | F1-score |
|---|---|---|---|---|
| YOLOv7 | 42.7 | 86.6 | 66.6 | 61.9 |
| YOLOv11 | 46.6 | 93.3 | 68.0 | 59.1 |
| **Proposed** | **73.8** | 79.4 | **76.5** | **76.9** |

Despite this domain shift, the proposed framework maintained competitive localization performance and achieved stronger specificity compared with several baseline methods. The relatively high specificity suggests that the proposed attention-guided filtering mechanism improved the ability of the network to suppress irrelevant activations under unseen imaging conditions. In contrast, models relying primarily on direct bounding-box regression appeared more sensitive to background variations.

However, sensitivity values remained lower across all models on the KUMC dataset, indicating that lesion localization under severe domain variation remains challenging. This behavior may be associated with differences in lesion appearance and image contrast between datasets. Since the proposed framework learns region-aware spatial representations derived from box-supervised learning, the localization response may become more conservative under unfamiliar visual patterns.

### *C. Cross-Dataset Evaluation on EDD2020*

The quantitative results obtained on the EDD2020 dataset are summarized in Table IV. The proposed framework achieved the strongest overall balance between sensitivity, specificity, and AUC among the evaluated models. In particular, the proposed method achieved an AUC of 0.881, outperforming YOLOv7 and remaining competitive with YOLOv11.

The improved performance observed on EDD2020 suggests that the multi-scale contextual representation improved robustness against lesion appearance variation. Small lesions and low-contrast regions often require simultaneous modeling of local texture and broader contextual information. By integrating multiple receptive fields within the bottleneck representation, the proposed framework was able to better preserve lesion evidence under challenging imaging conditions.

TABLE IV. POLYPS DETECTION (%) ON EDD2020 DATASET.

| Methods | Sensitivity | Specificity | AUC | F1-score |
|---|---|---|---|---|
| YOLOv7 | 78.0 | 76.6 | 77.3 | 85.0 |
| YOLOv11 | 74.0 | 96.7 | 85.3 | 84.7 |
| **Proposed** | **79.5** | 96.6 | **88.1** | **88.2** |

### *D. Cross-Dataset Evaluation on Kvasir-SEG*

Table V presents the quantitative evaluation results obtained on the Kvasir dataset. The proposed framework achieved the strongest overall cross-dataset performance, obtaining the highest AUC and F1-score among the compared YOLO-based detection methods. In contrast, YOLOv7 demonstrated extremely weak specificity under this dataset shift, indicating unstable discrimination between lesion and background regions.

TABLE V. POLYPS DETECTION (%) ON KVASIR-SEG DATASET.

| Methods | Sensitivity | Specificity | AUC | F1-score |
|---|---|---|---|---|
| YOLOv7 | 95.7 | 0.0 | 52.1 | 85.0 |
| YOLOv11 | 80.4 | 93.3 | 86.9 | 96.4 |
| **Proposed** | **96.7** | **93.3** | **95.0** | **98.2** |

The enhanced performance achieved by the proposed framework on Kvasir-SEG suggests that lesion-focused spatial learning improved robustness against dataset variability. Instead of relying solely on direct bounding-box regression, the proposed architecture first learns spatial lesion evidence and then converts these responses into localization outputs. This strategy appears to improve discrimination between true lesion structures and visually similar background patterns.

### *Overall Qualitative Comparisons*

Figure. 5 illustrates qualitative comparisons of activation responses and localization outputs obtained from YOLOv7, YOLOv11, and the proposed framework under challenging cross-dataset conditions. The first row shows an example from the KUMC dataset where all methods successfully localize the lesion region; however, the proposed framework produces a more compact and lesion-focused activation response with reduced activation in surrounding tissue regions. The second row corresponds to a challenging EDD2020 sample containing complex folds and illumination variations. In this case, YOLOv11 generates strong activations across large surrounding background regions rather than focusing only on the lesion area, whereas the proposed framework maintains more localized lesion-focused responses.

The third row illustrates a difficult example from the Kvasir dataset where YOLO-based detectors respond strongly to surrounding mucosal structures and broader anatomical regions, leading to less precise localization behavior and excessive background activation. In contrast, the proposed framework suppresses a substantial portion of these irrelevant responses and maintains stronger activation around lesion-related regions. The fourth row presents a low-contrast LDPolyp example with challenging illumination artifacts. In this case, YOLOv7 and YOLOv11 produce diffuse activations extending beyond the lesion boundary, while the proposed framework preserves a more localized response near the lesion region despite the difficult imaging conditions.

These qualitative results support the quantitative findings reported in Tables II–V, particularly the improved specificity and AUC values observed for the proposed framework. The heatmap responses indicate that the proposed method is less influenced by irrelevant background structures and produces more lesion-centered activations. This behavior helps explain its more stable performance under cross-dataset evaluation, where imaging appearance and background patterns differ from the training data. Therefore, the qualitative analysis provides visual evidence for the role of attention-guided feature suppression in improving detection-oriented localization robustness.

## V. CONCLUSION

The proposed method in this work introduced a robust polyp localization framework based on the design of a detection-oriented localization mechanism. Specifically, the localization framework adopted residual learning, attention-guided suppression, and context-aware representations at multiple scales. According to experimental evaluation conducted using LDPolyp, KUMC, EDD2020, and Kvasir datasets, the proposed localization mechanism outperformed YOLOv7 and YOLOv11 with regard to localization accuracy and cross-dataset robustness especially in difficult image capturing scenarios characterized by poor contrast ratios, complex tissue structures, and variations between data domains. Additionally, the qualitative results illustrated the robust nature of the proposed method for lesion region localization even in situations where the standard methods could not correctly identify lesion regions.

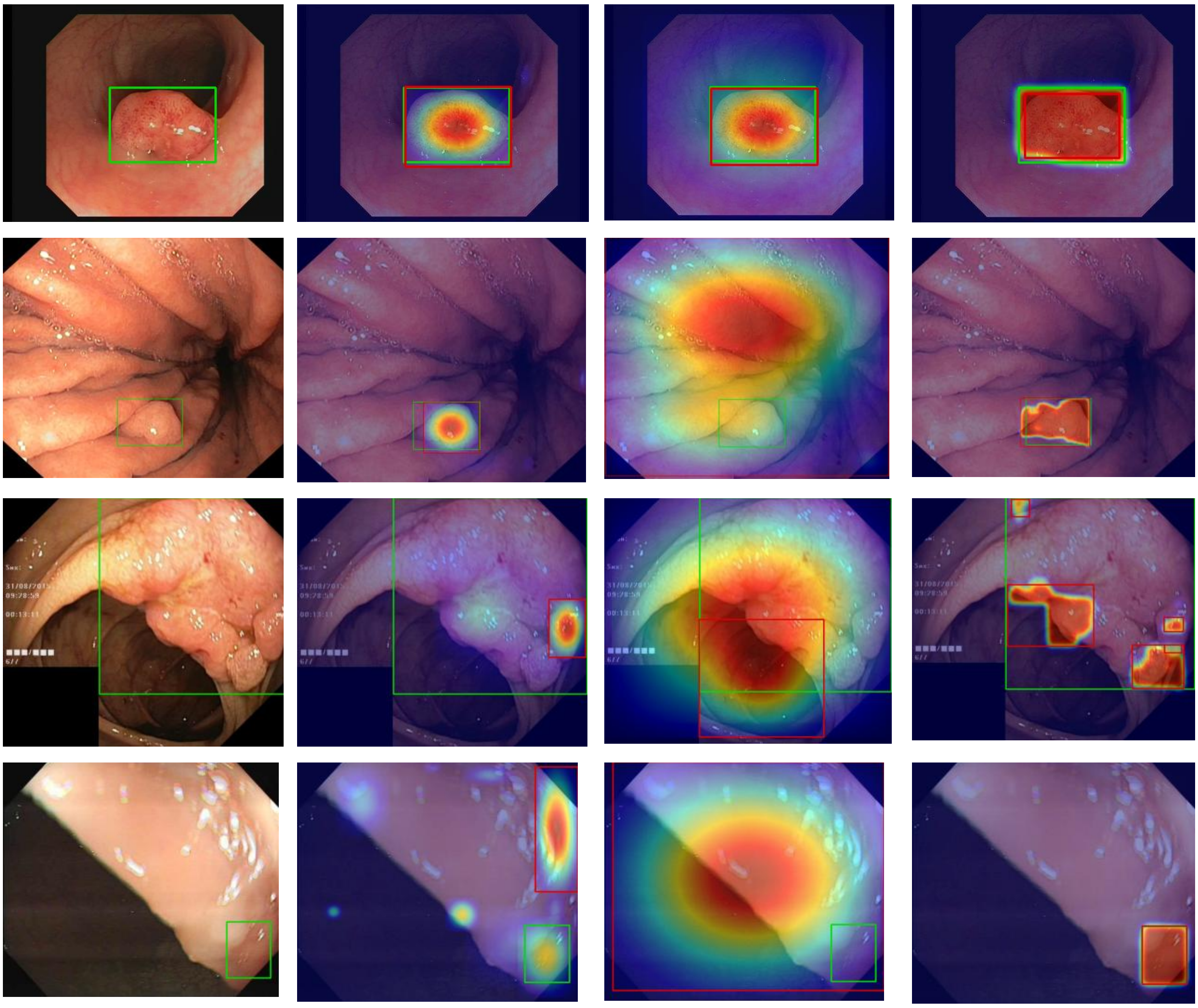

(a) (b) (c) (d)

Fig. 5. Qualitative comparison of activation responses and localization outputs for YOLOv7, YOLOv11, and the proposed framework under challenging cross-dataset conditions. rom top to bottom, the corresponding rows show representative samples from the KUMC, EDD2020, Kvasir-SEG, and LDPolyp datasets, respectively. (a) Ground truth lesion annotations. (b) YOLOv7 outputs. (c) YOLOv11 outputs. (d). Green boxes indicate ground truth lesion regions and red boxes indicate predicted localization outputs. The heatmaps illustrate the spatial activation behavior of each method during lesion localization.